\documentclass[lettersize,journal]{IEEEtran}
\usepackage{amsmath,amsfonts}
\usepackage{algorithmic}
\usepackage{algorithm}
\usepackage{array}
\usepackage[caption=false,font=normalsize,labelfont=sf,textfont=sf]{subfig}
\usepackage{textcomp}
\usepackage{stfloats}
\usepackage{url}
\usepackage{verbatim}
\usepackage{graphicx}
\usepackage{cite}
\begin{document}

\title{Arena-Bench: A Benchmarking Suite for Obstacle Avoidance Approaches in Highly Dynamic Environments}

\author{
\thanks{Manuscript received: February, 24, 2022; Revised May, 14, 2022; Accepted June, 12, 2022.}
\thanks{This letter was recommended for publication by Editor Tamim Asfour upon evaluation of the Associate Editor and Reviewers' comments.}
\thanks{$^{1}$The authors are with the Chair Industry Grade Networks and Clouds, Faculty of Electrical Engineering, and Computer Science,
		Berlin Institute of Technology, Berlin, Germany
		{\tt\small d.kaestner@tu-berlin.de}}Linh K{\"a}stner$^{1}$, Teham Bhuiyan$^{1}$, Tuan Anh Le$^{1}$, Elias Treis$^{1}$, Johannes Cox$^{1}$, Boris Meinardus$^{1}$,\\ Jacek Kmiecik$^{1}$, Reyk Carstens$^{1}$, Duc Pichel$^{1}$, Bassel Fatloun$^{1}$,  Niloufar Khorsandi$^{1}$ and Jens Lambrecht$^{1}$
\thanks{Digital Object Identifier (DOI): see top of this page.}
}

\markboth{IEEE Robotics and Automation Letters. Preprint Version. Accepted June, 2022}
{K{\"a}stner \MakeLowercase{\textit{et al.}}: Arena-Bench}  


\maketitle

\begin{abstract}
The ability to autonomously navigate safely, especially within dynamic environments, is paramount for mobile robotics. 
In recent years, DRL approaches have shown superior performance in dynamic obstacle avoidance. However, these learning-based approaches are often developed in specially designed simulation environments and are hard to test against conventional planning approaches. Furthermore, the integration and deployment of these approaches into real robotic platforms are not yet completely solved. In this paper, we present Arena-bench, a benchmark suite to train, test, and evaluate navigation planners on different robotic platforms within 3D environments. It provides tools to design and generate highly dynamic evaluation worlds, scenarios, and tasks for autonomous navigation and is fully integrated into the robot operating system.
To demonstrate the functionalities of our suite, we trained a DRL agent on our platform and compared it against a variety of existing different model-based and learning-based navigation approaches on a variety of relevant metrics. Finally, we deployed the approaches towards real robots and demonstrated the reproducibility of the results. The code is publicly available at github.com/ignc-research/arena-bench.
\end{abstract}


\section{Introduction}
\IEEEPARstart{M}{obile} robots are increasingly being employed for various use cases such as last-mile delivery, healthcare services, or operation in hazardous environments \cite{alatise2020review}.
Safe and reliable navigation in these highly complex and dynamic environments is essential for the operation of mobile robotics. In recent years, Deep Reinforcement Learning (DRL) has accomplished remarkable results for dynamic obstacle avoidance due to its ability to swiftly react to unexpected changes \cite{chiang2019learning},\cite{chen2017socially},\cite{faust2018prm}. A common barrier is that most of the research work evaluated their approaches on specifically designed simulation environments or test setups, making a general comparison against existing approaches difficult \cite{dugas2020navrep}. Furthermore, deployment and integration of DRL into real robotic platforms is still an open frontier due to safety reasons \cite{kiran2021deep}, \cite{zhu2021survey}. Thus, a benchmark to properly assess those approaches in realistic scenarios and against existing algorithms is not only an essential step towards the deployment of DRL into real robots but also assists in the development and validation of learning-based approaches on mobile robots.
Existing benchmarks for robot navigation algorithms mostly focus on static environments, but few exist that cover both dynamic and static ones. Moreover, existing benchmarks for navigation in dynamic environments often contain and compare only a small number of planners \cite{mrpb}, \cite{robobench}, \cite{rano2006steps}.
On that account, we propose Arena-bench, a benchmark suite consisting of tools to train, test, and evaluate navigation algorithms for dynamic obstacle avoidance on different robotic systems. This benchmark provides an intuitive interface to design and create dynamic scenarios within 2D and 3D simulators based on Flatland and Gazebo, respectively. To realistically simulate the environments, we integrated an extended version of Pedsim, which utilizes the social force model of Helbing et al. \cite{helbing1995social}. The benchmark is completely integrated into the robot operating system (ROS) and includes a variety of classic and state-of-the-art learning-based planners.
Arena-bench further provides tools to evaluate all planners on a variety of navigational metrics ranging from navigational safety and robustness to path quality and efficiency.
The platform also provides a set of different robots with various robot kinematics to test the approaches.

\begin{figure}[]
    \centering
    \includegraphics[width=0.47 \textwidth]{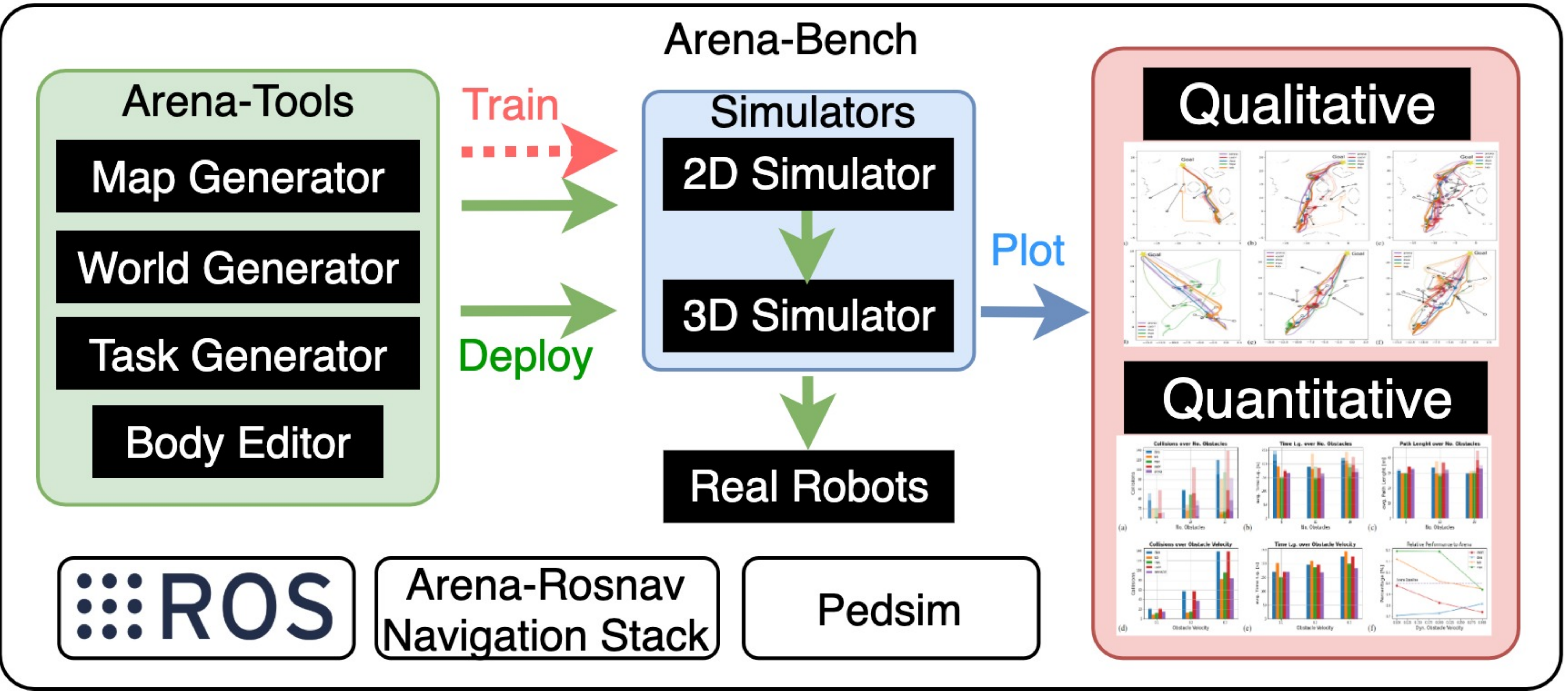}
    \caption{Arena-bench is a benchmark suite that enables to train and evaluate navigation approaches in realistic dynamic environments. It provides tools to develop navigation approaches, design and generate scenarios, and evaluation tasks on a variety of robotic platforms. The results can be plotted on a variety of different navigational metrics.}
    \label{intro}
\end{figure}

\noindent The main contributions of this work are the following:
\begin{itemize}
    \item Proposal of a benchmark suite to develop navigation approaches and design and generate realistic, dynamic scenarios. The Pedsim library is utilized to realistically model dynamic obstacles.
    \item Integration of a complete training pipeline to train DRL agents on different robots. The user can integrate new robots and train DRL algorithms on an efficient 2D simulator or on a more realistic 3D simulator.
    \item Extensive evaluation of several planners developed using this work and provision of tools to evaluate the results on a variety of relevant navigational metrics.
\end{itemize}

\section{Related works}
\noindent The fundamentals of this work are based on arena-rosnav proposed by Kästner et al. \cite{kastner2021arena}, which is a 2D simulator platform for testing, training, and benchmarking robotic navigation approaches in highly dynamic 2D environments. The results were previously presented in \cite{kastner2021arena}, where model-based, as well as learning-based methods, namely TEB \cite{teb}, DWA \cite{dwa}, MPC \cite{mpc}, CADRL \cite{cadrl}, and their DRL agent were evaluated and compared based on different performance metrics.
\\\noindent
A similar benchmark suite named Bench-mr was presented in a paper by Heiden et al. \cite{bench_mr}. It provides the utility to test, evaluate, and compare different motion-planning techniques in complex environment scenarios. Although the focus was placed on wheeled mobile robots, all notions of dynamic obstacles were disregarded. Thus, only static situations were considered. Other benchmarks that focus on static environments are presented in the works of Althoff et al. \cite{althoff2017commonroad} or Moll et al. \cite{moll2015benchmarking}. Nonetheless, path planning is a vital aspect of autonomous navigation, not only in static but also in dynamic environments. 
\\\noindent
Rano et al. \cite{rano2006steps} proposed a methodology used to evaluate the performance of different approaches, given that specific scenarios can vary from each other. Consequently, this methodology attempts to quantify several aspects of an actual environment, such as the amount of space occupied by obstacles and the space available for the robot to maneuver between different static obstructions. Based on these metrics, scenarios were classified and evaluated legitimately compared to previous methodologies.
\\\noindent
Robobench, introduced by Weisz et al. \cite{robobench} is another venture of a similar notion, trying to create uniform grounds for benchmarking in the field of robotics. They present a platform suitable for an array of tasks, such as navigation, manipulation, and observation. Different types of robots can be evaluated using an appropriate simulation engine and planning framework. The authors went a step beyond by taking the idea of virtualization and enabling users to save their simulations in a runnable state. Such containers can then be easily shared between users, making it easier for others to replicate results. Albeit functioning, with exemplary missions present to test out, the project did not grow much beyond its proof of concept.
\\\noindent
A different aspect of robotic navigation, namely Social Robot Navigation, was explored by Holtz et al. \cite{holtz2021socialgym}. The researchers developed a framework for benchmarking navigation approaches in dynamic human environments. This framework contributes an easily extensible 2D robot and human behavior simulator, an integration with OpenAI Gym interface \cite{brockman2016openai} for learning purposes, as well as an example benchmark scenario. Several learning policies were trained and evaluated in a novel environment. However, the authors only provided a comparison of learning-based approaches.

\begin{figure*}[!h]
    \centering
    \includegraphics[width=0.96\textwidth]{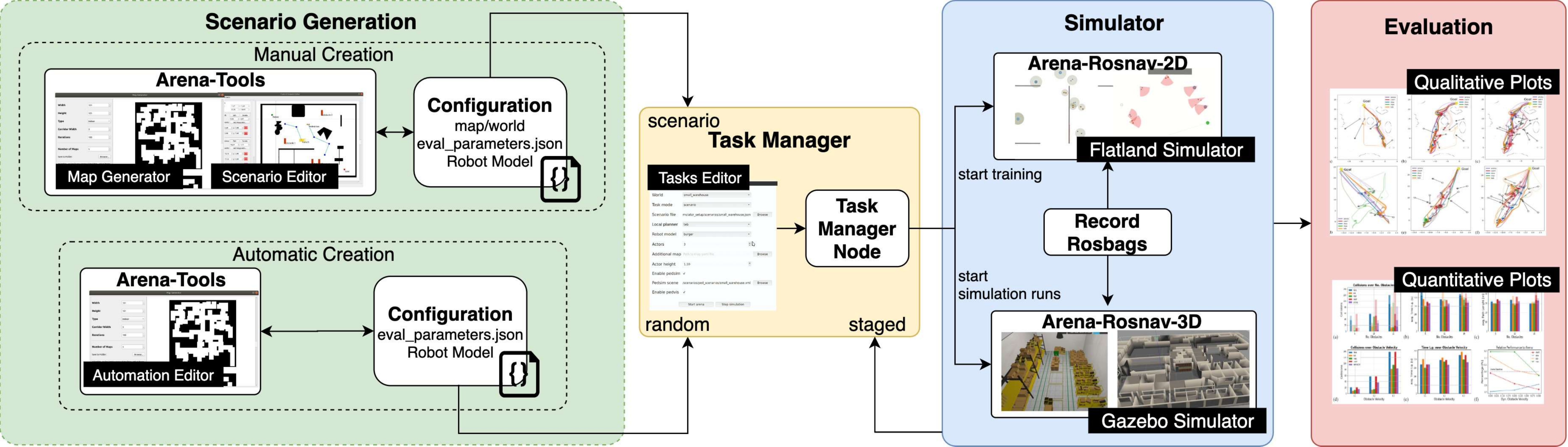}
    \caption{Our proposed benchmark suite consists of multiple modules for designing and generating different scenarios and evaluating different navigation approaches on specifically designed tasks. The suite includes a variety of different robot kinematics, including car-like, holonomic, and differential drive. The benchmark suite consists of a 2D and 3D simulation engine based on Flatland and Gazebo to conduct the evaluation runs. Finally, the evaluation class provides a variety of different metrics to visualize the test runs.}
    \label{system}
\end{figure*}
\noindent
Portugal et al. \cite{portugal2019ros} introduced \emph{patrolling\_sim}, a ROS and Stage-based \cite{stage} framework for simulation, benchmarking of multi-robot patrolling algorithms. The framework was packaged into an intuitive GUI, where the configuration of the simulation can be tuned to the user's needs. Additional tools were set up to integrate new algorithms, benchmark approaches, and compare the results. In particular, a node was established, which monitors the exchange of information in the system. Subsequently, the gathered data can be used to evaluate in terms of several performance metrics.
\\\noindent
In a recent paper by Wen et al., \cite{mrpb}, a similar ROS and Gazebo-based benchmarking platform under the name MRPB 1.0 was proposed. It allows researchers to compare their way of local planning against several existing state-of-the-art approaches. For the sake of this goal, diverse, dynamic, and complex scenarios were created. Dynamic obstacles were taken care of using Gazebo's actor principle by describing the desired looped trajectory. This way, several realistic, dynamic environments, e.g. an office and a shopping mall, were developed. Finally, two well-known local planners, DWA \cite{dwa}, and TEB \cite{teb} were evaluated on performance metrics such as smoothness, efficiency, and safety.
\\\noindent
Tsoi et al. \cite{tsoi2020sean} presented a simulation framework for social navigation is suggested by combining the features of ROS with state-of-the-art graphics and physics simulation of Unity. To facilitate the social navigation aspect of the platform, human models were introduced, with starting and goal position and behavior predicted according to the crowd flow presented in \cite{sohn2020laying}. Two different environments were supplied, a replica of an actual lab and a larger city scene. The evaluation toolkit allows for a systematical evaluation of navigation approaches by taking care of creating reproducible tasks and environments while collecting several performance metrics.
\\\noindent
All aforementioned benchmarks either focus on comparing classic model-based \cite{mrpb} or on learning-based approaches \cite{holtz2021socialgym}.
On the other hand, our benchmark is not limited to one paradigm and aims to provide a cross-discipline benchmark, enabling us to develop and evaluate both classic and learning-based navigation approaches directly on different kinds of mobile robots. We emphasize the development and deployment of DRL-based planners and provide a training pipeline to train these approaches.
Moreover, arena-benchmark provides an extensive set of tools to create and design unique evaluation tasks and scenarios.

\section{Methodology}
\noindent Our proposed benchmark consists of multiple modules that enable the user to design and generate evaluation scenarios, train DRL-based navigation algorithms, integrate ROS-based planners, and plot the results a variety of different navigational metrics.

\subsection{System Design}
\noindent The complete system design is illustrated in Figure \ref{system}. The first module in the benchmark suite is called arena-tools, which includes an extensive toolset to design and generate maps, scenarios, and tasks. The specific components are described in the next section.
The user can choose between a manual task generation mode consisting of generating their own map, specific scenarios, and an automatic generation of random tasks.
Using these tasks, the navigation approaches can be tested and benchmarked using the simulator module. Our framework is built on top of the 2D simulator Flatland and the 3D simulator Gazebo. Moreover, we provide the possibility to train DRL agents in both simulation environments. The resulting planners are cross-compatible across both simulators and utilize ROS as the middleware.
Finally, Arena-bench provides an automatic evaluation class to qualitatively and quantitatively evaluate all planning algorithms on up to 14 different navigation metrics ranging from navigational safety to trajectory quality and efficiency.

\subsection{Dynamic Obstacles with Pedsim}
\noindent The dynamic obstacles are spawned and controlled using this Pedsim library which, includes social states and calculates trajectories \cite{helbing1995social}. We integrated Pedsim into our 2D and 3D simulator for more realistic behavior and extended it with more behavior patterns and social states such as running, group talking, group listening, or waiting. Furthermore, we added a new service for changing the obstacles of the scene randomly and made use of the ability to spawn and re-spawn human agents during simulation. An exemplary scenario with multiple pedestrians is depicted in Fig. \ref{intro}.

\subsection{Map Generator}
\begin{figure*}[!h]
    \centering
    \includegraphics[width=0.95\textwidth]{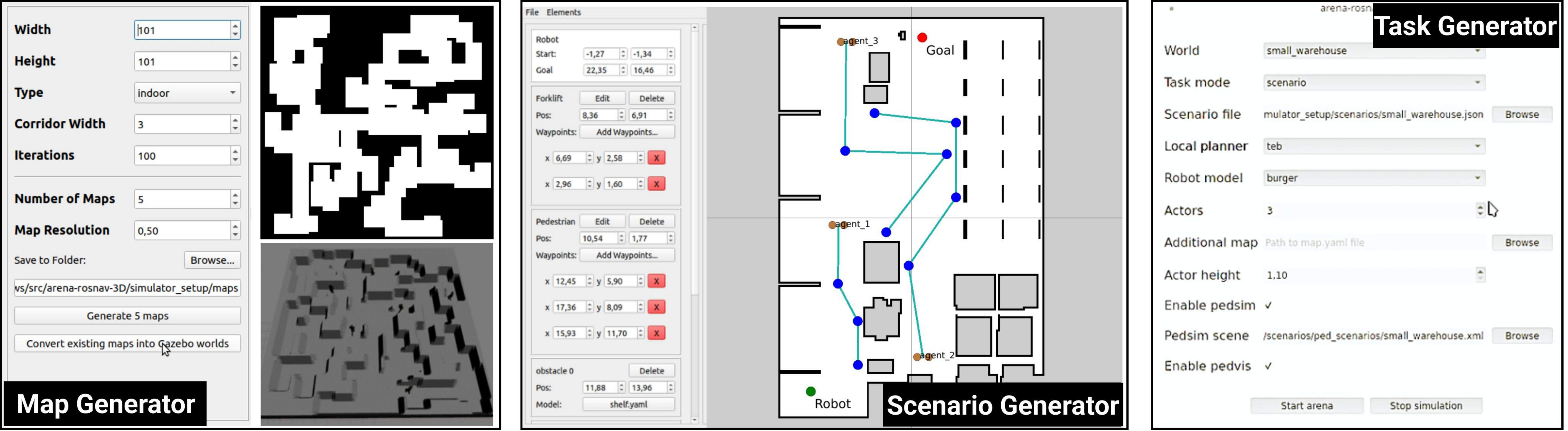}
    \caption{Arena-tools. We provide a variety of tools within our benchmark to generate or load existing maps, scenarios, and tasks. The detailed documentation of arena-tools can be found in the GitHub documentation. With the map editor, the user has the choice of loading an existing map or generating random maps with increasing difficulty. The user can choose between indoor and outdoor maps. The generated ROS maps can be transformed into a 3D Gazebo world. The scenario editor lets the user construct and design specific scenarios with specific start and endpoints as well as define the obstacle behavior. This includes their movements patterns, social states, and velocities. Using the task editor, the user can select parameters like planning algorithm, robot model, and task mode.}
    \label{guis}
\end{figure*}

\noindent Part of arena-tools is the map editor, which generates 2D and 3D worlds.
The user can either manually create a scenario with dynamic and static obstacles or select parameters for automatic scenario generation. Both can be done with our arena-tools application, which provides a variety of functionalities for scenario generation. Thus, even specific scenarios from the end-user can be constructed, which is crucial for real-world validations. Exemplary maps are illustrated in Fig. \ref{sim-results}.
We integrated the random map generation approach of Heiden et al. \cite{bench_mr} and extended it for generating 3D gazebo environments.
Moreover, we included the option to create increasing levels of difficulty by generating increasing amount and size of obstacles for outdoor maps and decreasing corridor size for indoor maps. The user has the option to randomly set obstacle positions and corridors after each episode (random mode) or use a specific curriculum (staged mode).

\subsection{Task Generator}
\noindent Once scenarios are generated, tasks within these scenarios can be defined using the task editor. Thereby, the user defines configuration files, which indicate the task mode, the number of runs, used planners, and the robot model. Fig. \ref{guis} illustrates the editor with all parameters to set. In total, the task generator has three modes: Random, Scenario, and Staged, which will be described in the following sections.
\\
\noindent \textbf{The Random Mode}
creates a random scenario for the used Gazebo world in each run. Different arbitrary locations are used for spawning the robot, as well as setting its desired goal position. We ensure to exclude setting these positions to the same location. The number of dynamic obstacles is specified before starting the simulation and fixed for the whole duration. The actors' starting and goal position is randomly generated based on the provided map. This information is then sent to the Pedsim simulator when creating the desired Pedsim obstacles.
The randomness of this mode makes it the preferred choice for quantitatively benchmarking different navigation approaches assuming that a large number of random scenarios are created and evaluated to ensure statistically significant evaluations.
\\
\noindent \textbf{The Scenario Mode} is a reasonable choice for both qualitative and quantitative evaluations. The idea is to limit the randomness without sacrificing complexity to deliver reproducible scenarios, which can be tested and evaluated using separate planning methods. Consequently, each run of this task mode should provide the same environment except for the dynamic behavior of the robot and human agents, which can not be accounted for. We achieve this through arena-tools \cite{tools}, where a scenario editor was developed. Existing maps of the environment can be used to then specify the robot's spawning location and goal position. Furthermore, we can add an arbitrary number of dynamic obstacles, describe their start/goal spot, as well as their waypoints, which should be included in their overall trajectory. The scenario editor generates a configuration file describing the specifics of a scenario. The data is read by the task\_manager, converted into a compatible format for our platform, and subsequently used in the simulation.
\\
\noindent \textbf{Staged Mode}
The staged mode is the preferred choice to automate a long list of evaluation scenarios. Here, the user specifies several levels with increasing difficulty and can set the threshold of when to reach this level. Thus, an evaluation curriculum is generated, which makes it an appropriate choice, especially for quantitative evaluations of planners. The stages can also be designed using arena-tools and are encoded into a configuration file.

\subsection{Navigation Stack and Planning Approaches}
\noindent Our system is fully integrated into ROS, which enables the usage of existing ROS functionalities. In our evaluation, a classic A-star approach by Hart et al. \cite{hart1968formal} was utilized as a global planner.
The local planning approaches are distinguished between model-based and learning-based approaches. While the model-based approaches work directly with the ROS move base interface, the learning-based approaches are trained using OpenAI Gym and require an additional entity to down-sample the global path due to the myopic nature of DRL approaches \cite{dugas2020navrep}. Therefore, we utilize an intermediate waypoint generator of our previous work \cite{9636039} to mitigate the myopic nature of DRL agents by providing a shorter goal horizon and avoiding local minima. The training pipeline and system design with the intermediate planner as an interconnection entity was presented in our previous work. The complete integration of classic move base planners and learning-based planners into ROS, enables them to be deployed and compared side-by-side, which makes direct comparison feasible.
Currently, three classic planning approaches MPC \cite{mpc}, TEB \cite{teb}, DWA \cite{dwa}, and four learning-based approaches, NAVREP \cite{dugas2020navrep}, Gring \cite{guldenring2020learning}, as well as our trained approach called ROSNAV are integrated. The ROSNAV planner was trained using the 2D simulator of our arena-bench suite for efficient training. It takes the structure and reward system of the agent presented by our previous work \cite{kastner2021arena} and was retrained with an adjusted action and observation space to match the specific robots. We utilize the following continuous action space $A$ for the different robots: Turtlebot3 (TB3) - $v^{TB3}$, Jackal (JKL) - $v^{JKL}$, and Robotino (RTO) - $v^{RTO}$:
\begin{align}
    a = \{v_{lin}& , v_{ang} \} \\
    v_{lin}^{TB3} \in [0, 0.22] m/s, \quad
    & v_{ang}^{TB3} \in [-2.7, 2.7] m/s \\
        v_{lin}^{JKL} \in [-2.0, 2.0] m/s, \quad
    & v_{ang}^{JKL} \in [-4.0, 4.0] m/s \\
        v_{lin,x,y}^{RTO} \in [-2.78, 2.78] m/s, \quad
    & v_{ang}^{RTO} \in [-1.0, 1.0] m/s
\end{align}

\noindent A complete list of robot parameters can be found on the arena-bench GitHub wiki page. Furthermore, we included our All-in-One planner (AIO) presented in our previous work \cite{kastner2021all}, which is a trained DRL-based control switch, which can choose between different planners based on specific situations.

\begin{figure*}[!h]
    \centering
    \includegraphics[width=0.99\textwidth]{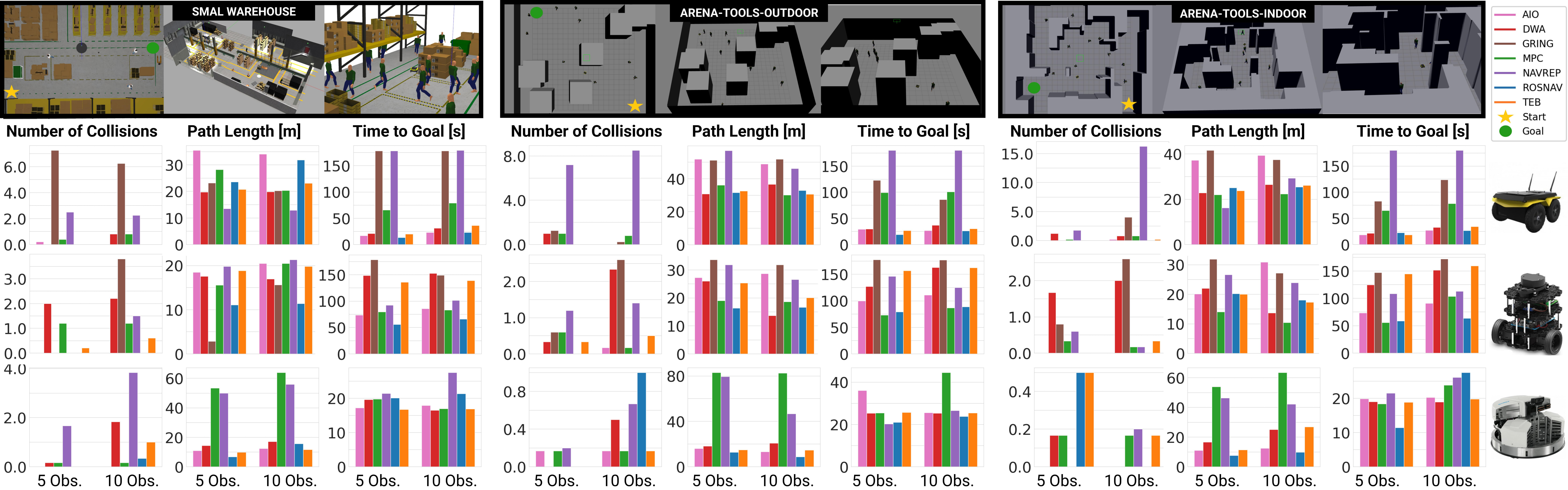}
    \caption{Experiments in simulation. Quantitative results of all planners on three robots over three worlds within the 3D simulation over the number of obstacles. The small warehouse world was integrated from a third-party package, and the outdoor and indoor environments generated using arena-tools. Collision rates, path length, and movement jerks are plotted to represent navigational safety, efficiency, and smoothness, respectively.}
    \label{sim-results}
\end{figure*}
\setlength{\dbltextfloatsep}{0.2cm}

\section{Evaluations}
\noindent To demonstrate the functionalities of our benchmark, we conducted experiments on several different maps, which were either imported from existing open source projects like the AWS-robomaker project or created using the arena-tools functionalities. For our evaluations, we included the small warehouse world from AWS-robomaker (20x20m) and an indoor and outdoor world created with our randomized world generator (15x15m).
Each world contains two scenarios with 5 and 10 pedestrians with an average speed of 0.3m/s. In each scenario, the obstacle velocities are set to 0.3 m/s. As a global planner, A-Star is used for all approaches. For the DRL-based planners, we utilized the Spatial Horizon waypoint generator of Kästner et al. \cite{9636039} using the time and location horizon of $t_{lim}=4s$ and $d_{ahead}=2m$. For the model-based approaches, we utilize the ROS move-base interface. Localization is acquired using the Adaptive Monte Carlo (AMCL) module. For each planner, we conduct 15 test runs on each scenario. During the test runs, our platform will record the necessary data automatically and provide it to our evaluation class, which is able to generate qualitative and quantitative plots on a variety of metrics listed in Table \ref{tablehyper}. It includes metrics to evaluate navigational efficiency, robustness, safety, and smoothness. Furthermore, some metrics are subdivided in sub-metrics such as the average, minimum, maximum and normalized values, which could give additional insight.

\begin{table}[!h]
\centering
	\setlength{\tabcolsep}{0.2pt}
	\renewcommand{\arraystretch}{0.5}
		\caption{Evaluation metrics}
	\begin{tabular}{l c r}
		\hline
		Metric  &Unit & Explanation    \\ \hline
		Success Rate$^{2}$ & \%        & Runs with < 2 collisions          \\ 
		Collision$^{1,2}$ & -        & Total number of collisions\\
		Time to reach goal$^{2}$& [$s$] & Time required to reach the goal   \\ 
		Path Length$^{1,2}$  & [$m$]    & Path length in m           \\ 
		Velocity (avg.)$^{2}$     & [$\frac{m}{s}$]  & Velocity of the robot \\
		Acceleration (avg.)$^{2}$ & [$\frac{m}{s^2}$] & Acceleration of the robot \\
		Movement Jerk$^{2}$ & [$\frac{m}{s^3}$] & Derivation of Acceleration \\ 
		Curvature(avg.,max.,min.,norm.)$^{2}$ & [$m$] &  Degree of trajectory changes\\
		Angle over length$^{2}$  & $[\frac{rad}{m}]$ & Curvature overthe  path-length \\
		Roughness$^{2}$ & -                 & Quantifies trajectory smoothness \\ 
		Clearing Dist.(avg.,max.,min.,norm.)$^{2}$ & $[m]$ & Distance kept to obstacles \\
		\hline
	\end{tabular}
    \footnotesize{$^1$Quantitative metric, $^2$Qualitative metric}\\
	\label{tablehyper}
	
\end{table}

\vspace*{-1mm}
\begin{figure*}[!h]
    \centering
    \includegraphics[width=0.95\textwidth]{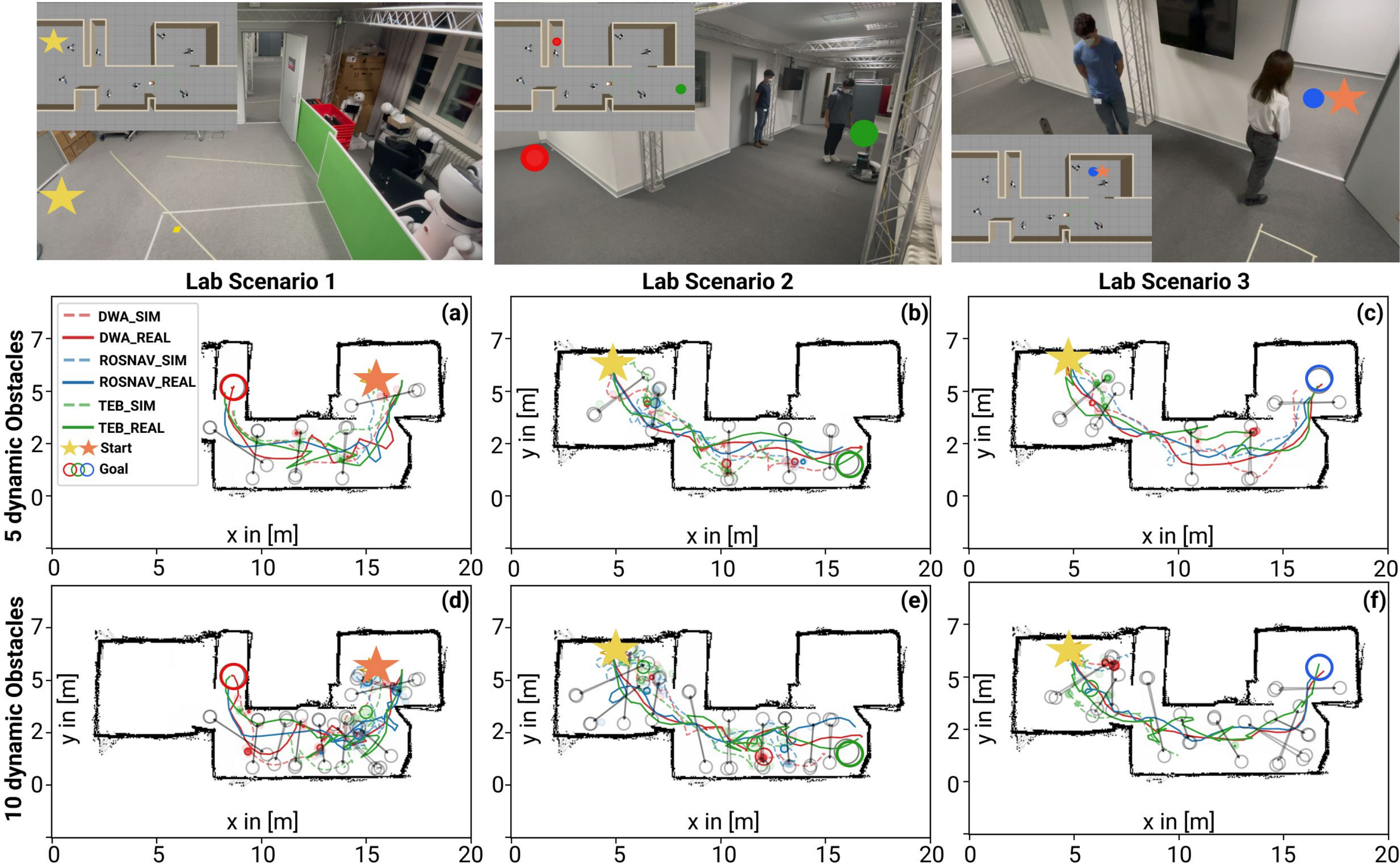}
    \caption{Experiments on the real robot. Upper row: Robotics laboratory where the real experiments were conducted and generated 3D world. The goals are illustrated as colored dots. The two different start positions are indicted by stars of different colors (yellow and orange). Lower row: qualitative plots of the real test runs. Goals are illustrated as colored circles to better visualize the fluctuation of the goal position due to non-optimal localization on the real robot. The runs on the real robot are bold, whereas simulation runs are dashed. The obstacle trajectories are illustrated in black. Out of the 15 different runs conducted, one exemplary trajectory is selected.}
    \label{real-results}
\end{figure*}

\subsection{Simulation Results}
\noindent  To represent navigational safety, efficiency, and smoothness, the aggregated collision rates, path lengths in meters, and movement jerks of all planners for three different robot platforms on each map are depicted in Fig. \ref{sim-results} respectively. Thereby, a collision is counted when the laser scan detects a value smaller than the robot radius. The movement jerk expresses the rate at which the robot changes its acceleration with respect to time. To represent all common robotic kinematic platforms, we deployed all planners except for the GRING planner on the Jackal (Ackerman Drive), the Turtlebot3 (TB3) (Differential Drive), and the Robotino (RTO) (Holonomic) robot. The GRING planner was only deployed on the Jackal and Turtlebot3 because it was only trained for wheeled robots, and deployment on the Robotino resulted in flawed behavior.
\\
\noindent \textbf{Navigational Safety:} It is observed that the AIO planner excels in terms of navigational safety, accomplishing the lowest collision rates in all scenarios with 5 and 10 obstacles. Our ROSNAV planner also accomplishes competitive results in terms of navigational safety with low collision rates. Only for the Robotino, ROSNAV produced collisions in the indoor and outdoor map. On all other robot platforms and maps, no collisions were produced. Based on the presented results, the classic move-base planners TEB, DWA, and MPC planners follow up in terms of navigational safety with competitive results over all robots and maps. In general, they also produce low collisions and high success rates over all scenarios and on all robots. However, only DWA on the Turlebot3 produces high collision rates. Aggravating factors are the slow maximum velocity of the Turlebot3 and computationally more demanding calculation times for DWA, which result in slower reaction times to incoming obstacles. The performance of the other learning-based approaches GRING and NAVREP is competitive in scenarios with 5 obstacles but drops significantly for scenarios with 10 obstacles. In situations with a high amount of dynamic obstacles, both planners are not able to react in time.
\\
\noindent \textbf{Navigational Efficiency:}
With regards to navigation efficiency, differences between the robot platforms can be observed. The AIO planner produces high path lengths for the Jackal and Turlebot3. However, for the Robotino, AIO is significantly more efficient, accomplishing one of the lowest path lengths compared to all planners. 
Whereas ROSNAV results are mediocre for the Jackal and Turlebot3, it outperforms all other planners on the Robotino, similar to our AIO planner. This indicates that our ROSNAV planner works best on holonomic platforms due to the more flexible set of movements the robot can exercise. The move-base planners TEB and DWA produce similar results with competitive results over all robots, maps, and scenarios. However, the MPC planner produces competitive results only on the Jackal and Turlebot3 but performs worst on the holonomic Robotino. Since MPC was specifically designed for car-like robots, these results are expected. 
GRING and NAVREP perform worst with regards to efficiency producing high path lengths in most scenarios and all robots. 

\noindent \textbf{Navigational Smoothness:} In terms of navigational smoothness, ROSNAV produces mediocre results with higher movement jerk values compared to the classic model-based planners DWA, MPC, and TEB, which all produce competitive results. Especially the MPC planner accomplishes the lowest movement jerk values on all robots and scenarios. Whereas on the Jackal and Robotino robots, the movement jerk of MPC is close to zero, the jerk is higher on the Turlebot3. Similar observations can be made for most of the other planners, which produce higher jerks for the Turlebot3. This is due to the low maximum velocity of the robot, which results in abrupt velocity changes that cause higher values. The findings also show that the movement jerk is directly correlated with the maximum velocity of the robot. Thus, the Robotino produces a significantly higher jerk compared to the Turlebot3 robot. Nevertheless, the MPC planner is still able to accomplish a  very low movement jerk on the Robotino and outperforms all other planners in terms of navigational smoothness.

\begin{figure*}[!h]
    \centering
    \includegraphics[width=0.95\textwidth]{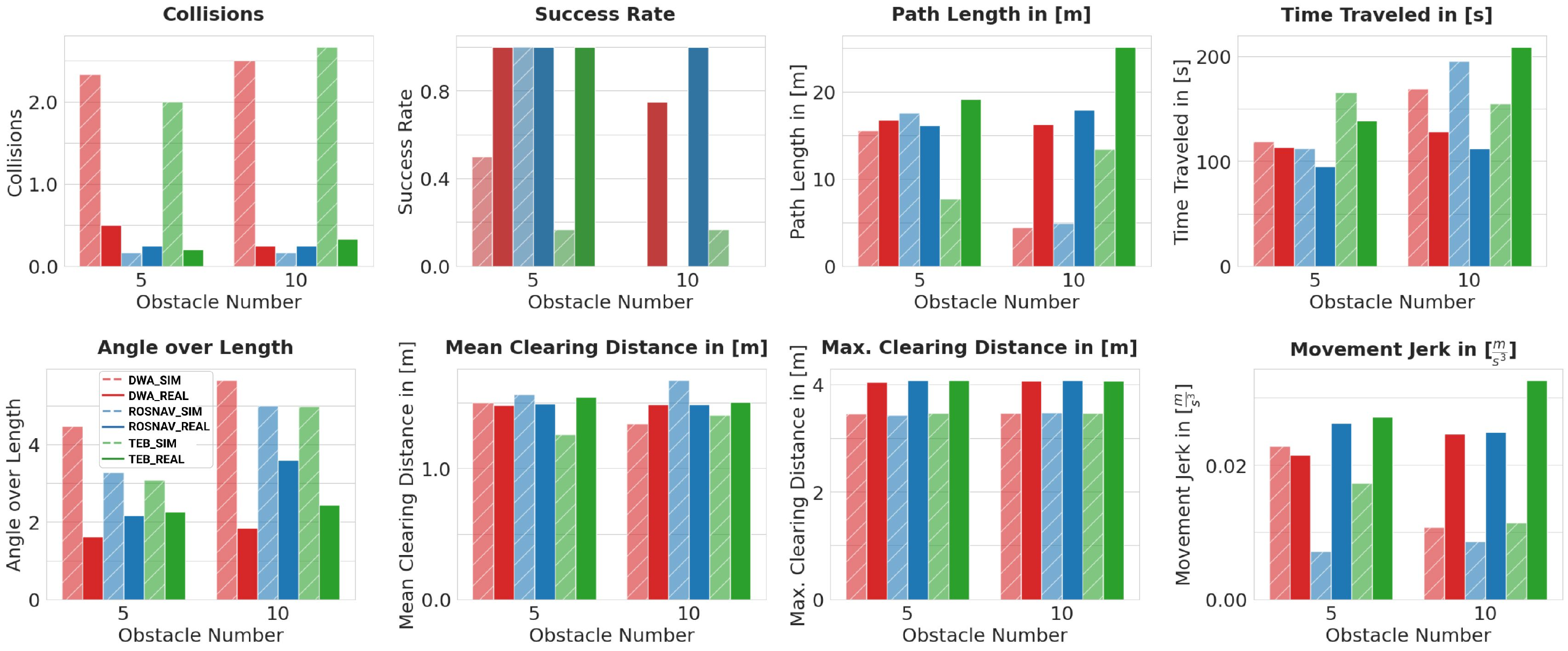}
    \caption{Quantitative plots for the real world tests. (Upper row): collisions, success rate, time traveled and path length indicate navigational efficiency and robustness, (Lower row): Angel over lengths, clearing distances, and movement jerks indicate navigational safety and smoothness. The metrics are plotted over an increasing number of obstacles.}
    \label{real-quanti}
\end{figure*}

\setlength{\dbltextfloatsep}{0.2cm}
\subsection{Test on the Real Robot}
To validate our results and provide a proof of concept, we transferred the algorithms towards real robots and conducted extensive experiments on the Turtlebot3 within our robotics laboratory. In parallel, we used arena-tools to generate a 3D world based on a 2D map of the laboratory (16x7m), which was recorded using SLAM. Subsequently, we generated the same scenarios using our scenario editor to ensure consistent evaluations across simulation and reality. Finally, we conducted 15 runs per planner in simulation and reality on each scenario. For the localization of the robot, we used the AMCL package in simulation and reality. It is to be noted that the real-world experiments are subject to inaccuracies due to inconsistent trajectories, velocities, and behavior of the human pedestrians. To mitigate those disturbances, we included trajectory marks and visual cues to mimic the scenarios as close as possible. However, a bias can not be completely avoided within real-world experiments.
\\
\noindent \textbf{Qualitative Evaluations:}
The qualitative results for both simulation and real runs are illustrated side-by-side in Fig. \ref{real-results}. It is observed that the trajectories of the real robot on all runs are similar to the simulation and that our simulation results can be reproduced within real environments. As expected, it is observed that the movement jerk is higher on real robots compared to the simulation, where the paths are more smooth. Especially for TEB and DWA, this can be observed. For TEB the typically stop and backwards movements once an obstacle is perceived is clearly visible e.g. in in Fig \ref{real-results} a) [x: 12, y:1], c) [x: 13, y:1.8], or d) [x: 15, y:1]. These backward movements are more pronounced on the real robot compared to the simulation, with the real robot driving a significantly longer path backward when perceiving an incoming obstacle, which leads to longer overall paths. This effect is also visible on the DWA planner, albeit not as distinct as on the TEB planner. Our ROSNAV planner behaves similar to the simulation and reaches the goal without collisions. Contrarily, our ROSNAV planner performs a stop and, if necessary small backward movement or immediately turns away from incoming obstacles. It can perceive incoming obstacles from approximately 1 m and react timely to avoid them. The performance of all planners is visualized in action in our supplementary video.
\noindent Another difference between simulation and reality is the non-optimal localization. Therefore, it is observed that the robot reaches the goal at slightly different goal positions, whereas the simulation results can keep a consistent goal position. However, this gap is only slight and can be mitigated by using a more accurate localizer, which uses multiple sensor information. \\
\noindent \textbf{Quantitative Evaluations:}
\noindent Additional quantitative evaluations to demonstrate these findings are illustrated in Fig. \ref{real-quanti}, where 8 different metrics are plotted. In addition to the previously introduced metrics presented in the simulation results, we plotted the mean and max clearing distance in meters to assess navigational safety, the angle over length, and the success rate. The plots validate the simulation-to-reality difference. This is most obvious with the path length and time to reach the goal of the TEB planner. On the real robot, the TEB planner often drives a long distance backward whenever an obstacle is perceived, which is exercised in simulation but to a much smaller extent. This increases the path length and time to reach the goal, which is significantly higher compared to ROSNAV and DWA. Regarding collision and success rates, our ROSNAV planner behaves similar to the simulation and can reproduce the results almost perfectly. However, the ROSNAV planner produces higher movement jerk values compared to the simulation, which was already visible in the qualitative plots. Nevertheless, this was also observed for DWA and TEB and expected due to fluctuating real-world factors like battery status and motor strength. In terms of path lengths, time traveled, and clearing distances in scenarios with 5 obstacles the simulation-to-reality gap is less distinct, with only slight differences between simulation and reality, which demonstrates the reproducibility of our experiments on real robots on these scenarios. However, in scenarios with 10 obstacles, the planners on the real robots required less time to reach the goal compared to the same scenarios in simulation. This could be due to the more cautious human behavior as well as non-consistent velocities of the humans within the real-world experiments. This also explains the large discrepancy between simulation and reality regarding the better performance of real robots in terms of collision and success rates.
\\\noindent \textbf{Discussion:}
The experiments run on the real robot provided additional insight into the sim-2-real gap, which is valuable to validate reproducibility on real robots, optimize planners, or address aspects of the simulation. Despite the gap between simulation and reality on a number of metrics, the differences were expected and due to known factors like non-perfect localization, sensor noise, fluctuating real-world factors like battery and motor status, non-optimal floor areas, or human behavior.
However, the real-world evaluations could still validate the comparability of all planners in terms of different relevant metrics. Since the evaluations took place in Gazebo, which realistically simulates robot kinematics as well as sensor data, transfer towards real robotic platforms can be achieved without major issues assuming the observation and action space is consistent. Using the platform to compare different robotic systems and navigational metrics against each other provides valuable information to be considered and assessed when choosing an appropriate robot for specific scenarios. 
The findings of the quantitative evaluations assisted in the assessment of each planner, scenario, and robot. Whereas classic model-based planners like MPC, TEB, or DWA excel with regard to navigational smoothness, they struggle in highly dynamic environments, where learning-based approaches are a better choice. Furthermore, planners like MPC accomplish better results on wheeled robots like the Jackal, while our learning-based DRL agents can take advantage of the more flexible holonomic robot platform. A downside of our approach is that it needs to be trained on each robotic platform again. On the other hand, model-based approaches like MPC and TEB require finetuning of hyperparameters for each robot, which is tedious and not always trivial.

\section{Conclusion}
\noindent In this paper, we proposed Arena-bench, a benchmarking suite to train, test, and compare navigation approaches within ROS in highly dynamic simulation- and real-world environments. Therefore, we introduced a set of tools to design and generate worlds, scenarios, and tasks. Our platform enables the development and training of DRL approaches as well as the integration of classic planners for different robots. To demonstrate the functionalities, we included three different robot types, trained DRL agents on them, and included three conventional planners. Subsequently, we created a variety of test scenarios and conducted extensive evaluations of all approaches with a wide range of alternative navigation approaches. Finally, we transfer the approaches towards real robots and conduct field experiments to assess the sim-to-real gap. The platform provides an important tool not only to assist researchers in developing and evaluating navigation approaches in dynamic environments but also in deploying DRL-based planners towards real robots. In future works, we aspire to do more extensive evaluations on real robotic systems and compare the impact of different robot kinematics on the performance of the navigation approaches. Furthermore, we aim to elucidate the impact of different simulation environments for training and testing.


\typeout{}
\bibliographystyle{IEEEtran}
\bibliography{main}

\vfill

\end{document}